# A Rapid Instrument Exchange System for Humanoid Robots in Minimally Invasive Surgery

Bingcong Zhang[a], Yihang Lyu[a], Lianbo Ma[a], Yushi He[a], Pengfei Wei[a], Xingchi Liu[a], Jinhua Li[a,b,*], Jianchang Zhao[a,b,*], Lizhi Pan[a,b,*]

[a] Tianjin University, Tianjin, China

[b] Institute of Medical Robotics and Intelligent Systems, Tianjin University, Tianjin, China

*Corresponding authors. E-mail addresses lijinhua@tju.edu.cn;
zhaojianchang@tju.edu.cn; melzpan@tju.edu.cn

**Abstract**

Humanoid robot technologies have demonstrated immense potential for minimally invasive surgery (MIS). Unlike dedicated multi-arm surgical platforms, the inherent dual-arm configuration of humanoid robots necessitates an efficient instrument exchange capability to perform complex procedures, mimicking the natural workflow where surgeons manually switch instruments. To address this, this paper proposes an immersive teleoperated rapid instrument exchange system. The system utilizes a low-latency mechanism based on single-axis compliant docking and environmental constraint release. Integrated with real-time first-person view (FPV) perception via a head-mounted display (HMD), this framework significantly reduces operational complexity and cognitive load during the docking process. Comparative evaluations between experts and novices demonstrate high operational robustness and a rapidly converging learning curve; novice performance in instrument attachment and detachment improved substantially after brief training. While long-distance spatial alignment still presents challenges in time cost and collaborative stability, this study successfully validates the technical feasibility of humanoid robots executing stable instrument exchanges within constrained clinical environments.

## 1. Introduction

In recent years, with the deep integration of robotic technology into MIS, surgical robots have demonstrated irreplaceable advantages in navigating complex anatomical structures, enhancing operational precision, and alleviating surgeon fatigue [1-3]. However, mainstream surgical robot platforms widely used in clinical practice today are still hampered by several limitations, including high acquisition and maintenance costs, large system footprints, and a heavy reliance on specialized operating room layouts and steep learning curves [4-5]. In contrast, the rapid advancement of general-purpose humanoid robotics presents a unique opportunity to overcome these hardware barriers and facilitate more flexible surgical assistance

[6]. For example, Cho et al. pioneered the exploration of humanoid robots as first assistants in clinical procedures [7], while Liang et al. further demonstrated the vast potential of teleoperated humanoid robots in performing handheld laparoscopic tasks [6].

During complex minimally invasive surgical procedures, operational steps such as grasping, cutting, electrocoagulation, and suturing frequently necessitate the switching of various specialized surgical instruments. To accommodate this multi-instrument coordination, traditional surgical platforms are typically engineered as domain-specific systems or rely on redundant multi-arm configurations to host various tools simultaneously [8]. This hardware-stacking approach not only multiplies the equipment procurement costs for medical institutions but also further exacerbates the spatial congestion within already confined operating rooms. Given that humanoid robots are inherently constrained by their anthropomorphic dual-arm structure, equipping them with an efficient instrument exchange capability is not merely a necessity to compensate for the lack of redundant manipulators; it is a critical prerequisite for maximizing their high biomimetic dexterity and minimal spatial footprint. If a humanoid robot were to rely entirely on bedside assistants for manual instrument insertion and extraction during surgery, it would not only incur unnecessary labor and time costs but also severely disrupt the continuity of the surgical workflow, thereby significantly heightening the collaborative cognitive load between the primary surgeon and the assistant. Therefore, while the rapid exchange of instruments is not a core action that directly treats the lesion, it profoundly dictates the overall efficiency of the surgery, the clinical usability of the robotic system, and ultimately, surgical safety at a fundamental system level [9].

To optimize intraoperative workflows and reduce reliance on manual assistance, researchers have recently begun exploring rapid exchange interfaces and automated exchange mechanisms for surgical robots [10-11]. Although these solutions help enhance system modularity, most existing technologies are tailored for traditional multi-arm or dedicated medical robot configurations, making them difficult to directly integrate into humanoid robot platforms subject to a high degree of biomimetics and dual-arm spatial motion constraints. Furthermore, until fully autonomous instrument exchange technologies mature, teleoperation remains the most reliable approach for humanoid robots to execute high-precision instrument docking [12]. Motivated by this background and the associated practical challenges, this paper proposes an immersive teleoperated instrument rapid exchange system tailored for humanoid robots in minimally invasive surgery. By utilizing an HMD to present real-time visual feedback from the FPV, combined with natural three-dimensional spatial motion mapping and a rapid exchange architecture customized specifically for humanoid robots, this system aims to deliver a teleoperated instrument exchange solution that balances an intuitive operational experience, stable connections, and high fault tolerance.

## 2.Related works

## 2.1 Traditional Tool and Instrument Replacement Methods in Industrial and Medical Robots

In the mature fields of industrial and automated manufacturing, Automatic Tool Changers (ATCs) serve as core components for expanding the flexible manipulation capabilities of robots, backed by profound technological accumulation. Traditional mechanical gripper structures, such as cam-actuated ball-locking mechanisms or those featuring integrated micro-motors, eliminate the reliance on pneumatic pipelines and provide extremely high physical repeatability [13-14]. However, in high-precision operational scenarios with severely constrained spaces, such as autonomous surgery, instrument replacement often relies on human assistants; this is not only inefficient but also prone to introducing uncoordinated errors among medical staff. To address this clinical pain point, Wang et al. developed a novel minimally invasive surgical robot integrated with a quick-exchange interface. By utilizing multi-stage cable transmissions and a distal self-locking structure, they successfully compressed the physical exchange time of surgical instruments to under 20 seconds [15]. Similarly, Kim et al. proposed a semi-autonomous dual-arm robotic system for ear, nose, and throat (ENT) clinics. Hardware-wise, this system is equipped with a dedicated ToolBot to handle the automatic switching of various instruments, thereby achieving automated instrument changing for the robotic arms [16]. Nevertheless, such automated rapid-exchange solutions are highly dependent on their custom-tailored, specialized robotic architectures. Their specific interface designs and system volumes make them difficult to directly transfer to humanoid robot platforms, which feature high biomimetics and strict spatial motion constraints.

## 2.2 Research Progress on Specialized Rapid Exchange Mechanisms for Humanoid Robots and High-Precision Autonomous Systems

Although rich experience has been accumulated in rapid exchange devices for industrial and traditional multi-arm medical robots, their application in the field of humanoid robots remains almost entirely unexplored. Currently, the end-effector designs of humanoid robots are predominantly general-purpose grippers or biomimetic dexterous hands, aimed at meeting the grasping needs of daily objects. Consequently, related research on rapid exchange mechanisms primarily focuses on the exploration of lightweight structures and the application of smart actuation methods. For instance, Park et al. developed a retractable ratchet locking system driven by shape memory alloys (SMAs). Utilizing the physical interference between the ratchet and the pawl, this system achieves purely mechanical self-locking after a single contraction of the SMA, enabling zero-power holding during the maintenance phase [17]. Lee et al. further applied the concept of bistable mechanical structures to high-payload robotic grippers. By integrating flexible SMA actuators with a band-shaped self-locking closure ring, they achieved a physical absolute locking force of up to 7.85 kg with a self-weight of only 265 g [18]. Although these lightweight mechanisms based on smart materials exhibit massive potential in reducing energy consumption, their core designs remain confined to terminal grasping tasks. Faced with the requirements of high-precision docking of slender instruments, complex power transmission, and high-frequency switching demanded in minimally invasive surgery, existing gripper-based quick-change solutions exhibit a significant technological gap and completely fail to match the operational characteristics of specialized surgical instruments.

## 3. Methods

To address the previously discussed limitations regarding the integration of existing rapid exchange interfaces from dedicated surgical robots into humanoid platforms, as well as the current immaturity of fully autonomous operations for complex, delicate tasks, this section details the design and implementation of the proposed immersive teleoperated instrument rapid exchange system for minimally invasive surgery. The core architecture of this system is designed to maximize the spatial operational flexibility and dexterity of the humanoid robot via master-slave collaborative control.

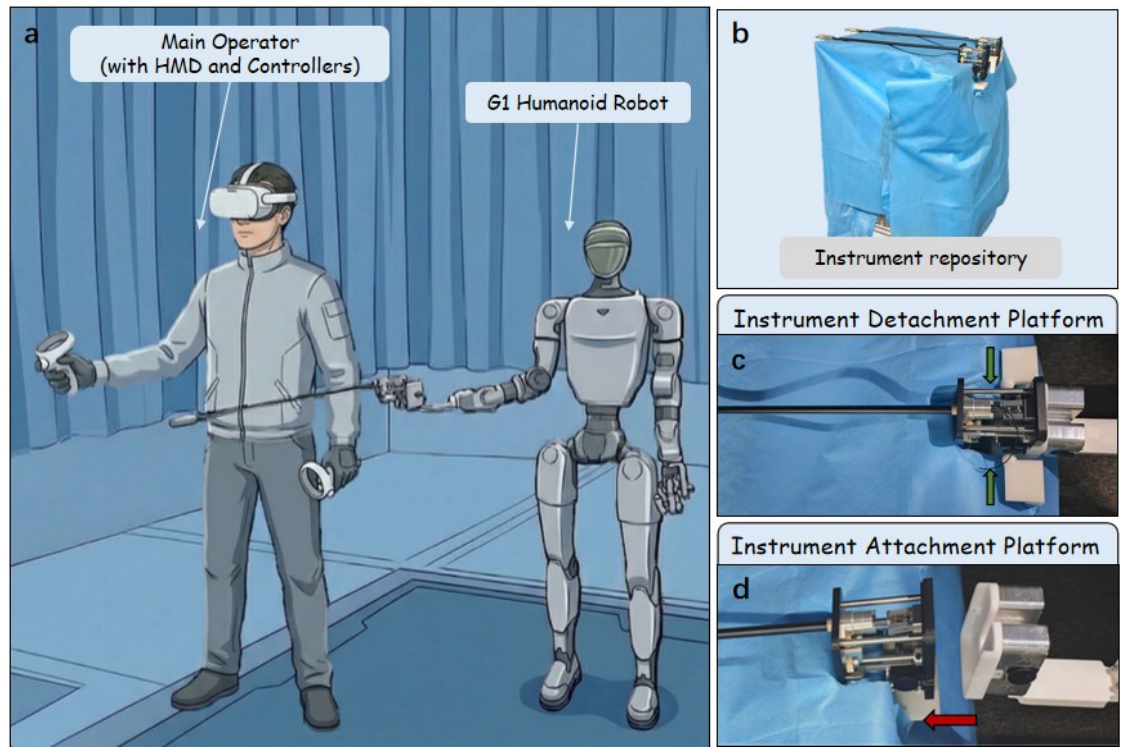

Figure 1. Overview of the immersive teleoperated surgical instrument rapid exchange system (a) This master-slave control architecture illustrates the main operator executing remote control by wearing an HMD and holding controllers, while the G1 humanoid robot acts as the slave execution platform. (b) The custom surgical instrument repository deployed within the robot's effective workspace. (c) The physical mechanism on the instrument detachment platform, demonstrating the release of the locking device during the instrument detachment phase. (d) The physical mechanism on the instrument attachment platform, illustrating the rigid terminal locking during the instrument attachment phase.

**3.1 Surgical Instrument Rapid Exchange Mechanism**

This study employs the G1 humanoid robot from Unitree Robotics as the slave execution platform within the master-slave control architecture. The robot features a total of 29 degrees of freedom (DoFs) across its full body, with 7 DoFs allocated to each of the upper and lower limbs bilaterally, and an independent yaw DoF at the waist. This high-DoF configuration provides ample kinematic redundancy for fine pose adjustments within constrained spaces.

To achieve efficient surgical instrument exchange, this study specifically designed a rapid exchange system and an accompanying instrument repository tailored to the humanoid robot's end effectors. The system primarily comprises two core structures: the robot-side active component and the instrument-side passive component. The active component is rigidly mounted to the robot's arm terminal and embeds a driving motor, providing the underlying actuation for core surgical actions such as grasper closure. The passive component is integrated at the base of the surgical instrument, utilizing a wire-driven mechanism to dock with the active component via a precise mechanical interface and receive power transmission.

The accompanying surgical instrument repository is independently deployed within the robot's effective workspace; its upper section is equipped with two dedicated docking bays to alternately accommodate the storage and exchange of instruments. The entire rapid exchange workflow relies on the synergistic operation of a terminal auto-locking mechanism triggered by direct axial insertion and a repository-assisted unlocking sequence.

During the instrument attachment phase, the system requires the active component on the robotic arm's terminal to precisely align with the tail interface of the pending instrument. As the robotic arm feeds forward axially, the end faces of the active and passive components fully engage, smoothly triggering the internal mechanical latch, thereby accomplishing the docking of the power transmission interface and the rigid terminal locking in a single continuous motion.

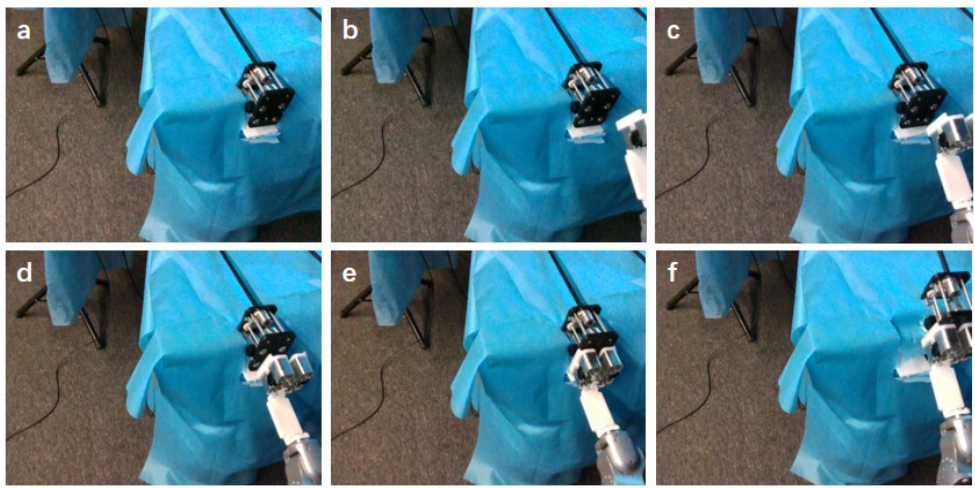

Figure 2. Detailed operational sequence of the surgical instrument attachment process. (a) Initial first-person view (FPV) from the humanoid robot's perspective before the task. (b-d) Spatial pose alignment process, where the operator teleoperates the robotic arm to align the active component

with the instrument's passive interface. (e) Axial docking and engagement of the rigid terminal self-locking mechanism. (f) Post-installation phase, illustrating the robotic arm withdrawing from the instrument repository while securely carrying the attached instrument.

During the instrument detachment phase, the currently mounted instrument must be smoothly returned to an empty bay in the repository. When the passive component of the instrument enters the docking slot and triggers the limit switch at the bottom, the auxiliary release mechanism integrated into the repository side is immediately activated, applying a pressing force to the unlock button on the instrument side. Assuming the unlocking force exerted by the release mechanism is $F_{\text{release}}$, to ensure a successful unlocking action, this force must overcome the inherent preload of the latch and frictional resistance; the mechanical constraint condition is:

$$F_{\text{release}} \geq F_{\text{lock,preload}} + c_{\text{fric}} \cdot F_{\text{normal}}, \quad (1)$$

where $F_{\text{lock,preload}}$ is the latch preload force, $c_{\text{fric}}$ is the friction coefficient during the unlocking process, and $F_{\text{normal}}$ is the normal force at the latch contact surface. Only when this unlocking condition is met will the rigid mechanical constraints within the system be released. In this state, the robotic arm retracts linearly along the axis. At this point, the separation resistance between the active and passive components, denoted as $F_{\text{withdraw}}$, drops precipitously, and its calculation model transitions to comprise solely the residual contact force and sliding friction:

$$F_{\text{withdraw}} \geq F_{\text{residual}} + \mu_{\text{interface}} \cdot N_{\text{interface}}, \quad (2)$$

where $\mu_{\text{interface}}$ is the kinetic friction coefficient of the interface joint surface, and $N_{\text{interface}}$ is the normal pressure on the joint surface. Because the rapid exchange button is effectively pressed to achieve mechanical unlocking, the locking force that originally impeded axial movement is eliminated. Based on this collaborative mechanism, the robotic arm only needs to exert a relatively small pulling force to overcome residual friction, achieving a smooth separation of the terminal active component from the surgical instrument.

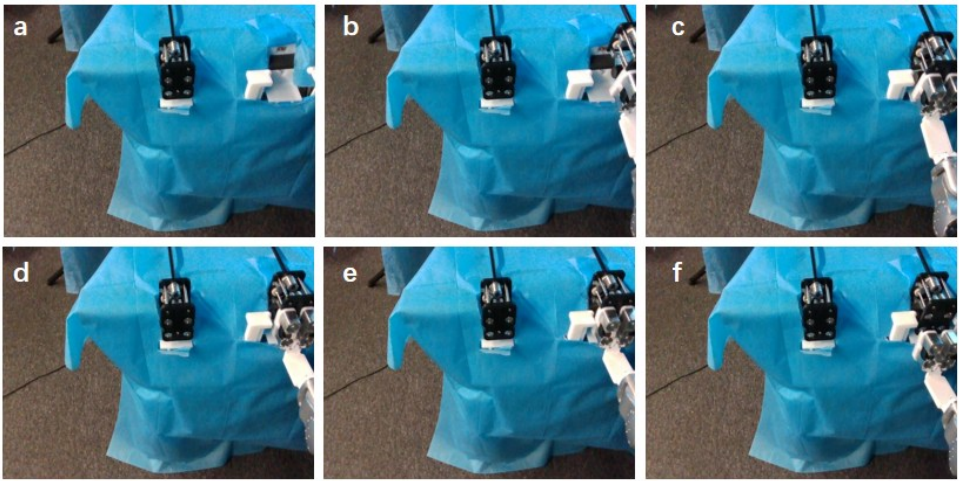

Figure 3. Detailed operational sequence of the surgical instrument detachment process. (a) Initial first-person view (FPV) from the humanoid robot's perspective before the task, with the robotic arm holding the surgical instrument. (b-c) Spatial pose alignment process, where the operator teleoperates the robotic arm to align the instrument with an empty docking slot in the repository. (d) Axial insertion and triggering of the auxiliary release mechanism. (e) Linear axial withdrawal of the terminal active component. (f) Post-detachment phase, illustrating the robotic arm completely separated from the safely stored instrument.

**3.2 Immersive Teleoperation and Visual Feedback Framework**

This system adopts an immersive teleoperation paradigm based on a master-slave architecture. The operator executes remote master-slave control by wearing an HMD (model: PICO 4 Enterprise) and holding spatial controllers. The system streams the FPV captured by the robot's head-mounted visual sensors to the HMD terminal with low latency, providing the operator with situated visual perception. Simultaneously, the spatial pose commands from the operator's handheld controllers are mapped in real-time as kinematic inputs to the robotic arms, thereby realizing intuitive three-dimensional spatial hand-eye coordinated control.

**3.3 Experimental Design and Evaluation Protocol**

Evaluating an immersive teleoperation-based instrument rapid exchange system necessitates not only verifying the physical reliability of the underlying mechanical architecture but also assessing the operator's hand-eye coordination efficiency and cognitive load under this control modality. Specifically, time efficiency and operational success rate can directly reflect the effectiveness and fault tolerance of the rapid exchange mechanism's physical design. Meanwhile, comparative performances among subjects with varying experience levels, alongside the convergence rate of their learning curves, can objectively quantify the intuitiveness of this approach, thereby evaluating the system's learning threshold and ergonomic performance in human-robot interaction. Based on this evaluation logic, this study designed a rigorous, grouped comparative experiment.

Subjects were divided into two groups based on their operational experience: researchers with extensive experience in teleoperating humanoid robots constituted the Expert Group, while personnel with no prior operation experience formed the Novice Group.

In the specific rapid exchange performance evaluation experiments, subjects were required to wear the HMD and hold the controllers to complete standardized attachment and detachment tasks via immersive teleoperation. Time recording for the entire procedure commenced the moment the operator issued the first movement command. In the attachment task, the operator relied on the robot's first-person vision transmitted through the HMD to guide the robotic arm terminal above the instrument repository and fine-tune its pose for alignment. Subsequently, the operator controlled the robotic arm to push forward axially until the internal mechanical latch was triggered to complete the final locking. In the detachment task, the operator again utilized 3D visual feedback to teleoperate the robotic arm, precisely pushing the used instrument into an empty docking bay. Upon observing the activation of the repository's auxiliary release mechanism, the operator commanded the robotic arm to retract linearly to achieve a smooth unloading.

The experimental tasks were decomposed into three standardized sub-modules: single instrument attachment, single instrument detachment, and a complete exchange cycle. The comprehensive performance of the system was quantitatively evaluated through the following core metrics:

1) Time Efficiency: The average completion time for each task module, recorded from the initial command to complete detachment. Let the total time for a single instrument exchange be $T_{exchange}$, which comprises the detachment time $T_{unload}$ and the attachment time $T_{install}$:

$$T_{exchange} = T_{unload} + T_{install}. \tag{3}$$

Based on this workflow, the time consumed in these phases can be further detailed as the sum of the times for the respective action sub-modules:

$$T_{unload} = T_{move,return} + T_{trigger,release} + T_{withdraw}, \tag{4}$$

$$T_{install} = T_{align} + T_{feed} + T_{lock}, \tag{5}$$

where $T_{move,return}$, $T_{trigger,release}$, and $T_{withdraw}$ correspond to the time required for moving back, triggering the release, and axial retraction during the detachment process, respectively. $T_{align}$, $T_{feed}$, and $T_{lock}$ correspond to the time required for pose alignment, axial feeding, and rigid locking during the attachment process, respectively.

2) System Robustness: The operational success rate of successfully completing the rapid exchange docking on the first attempt across multiple repeated trials. The instrument exchange success rate $P_{success}$ is further defined as:

$$P_{success} = \left(1 - \frac{n_{fail}}{n_{total}}\right) \times 100\%, \tag{6}$$

where $n_{fail}$ is the number of unsuccessful trials resulting from alignment deviations, un-triggered latches, or unlocking failures, and $n_{total}$ is the total number of planned exchanges.

3) Learning Curve: Tracking the number of training rounds required for novice subjects to reach the expert group's average time baseline, aiming to assess the system's learning threshold and operational intuitiveness.

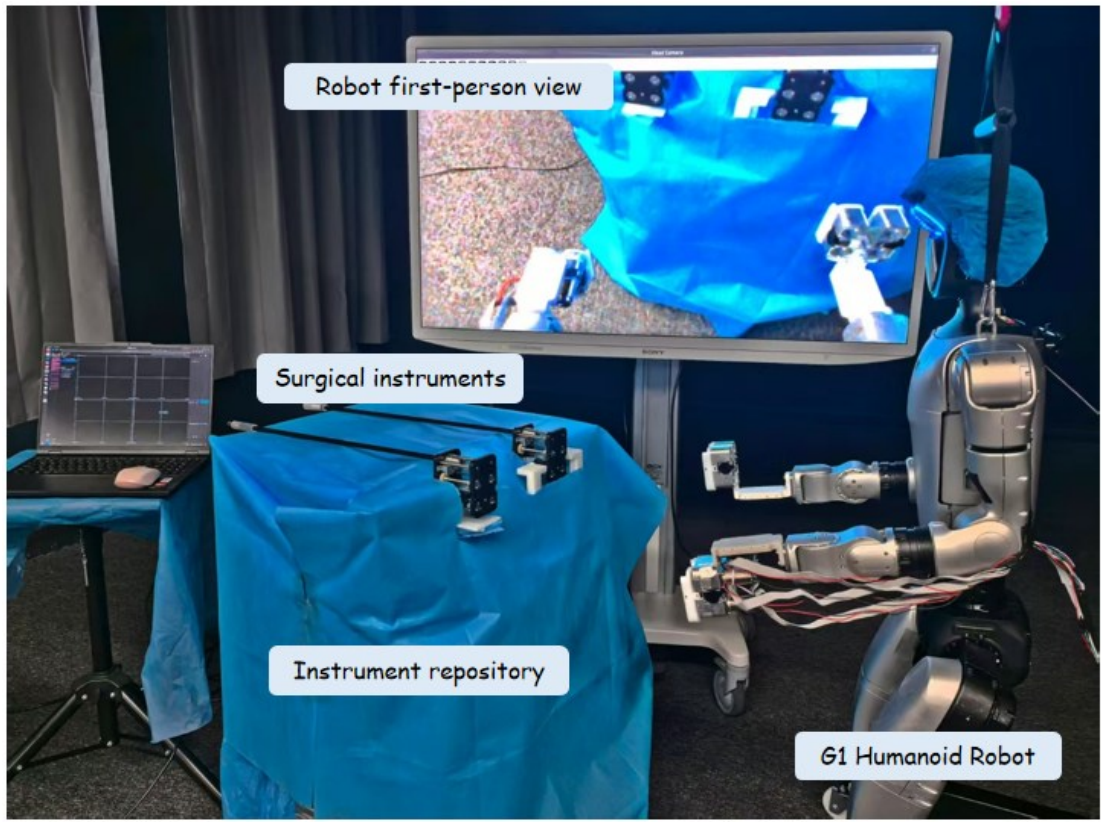

Figure 4. Experimental setup for evaluating the surgical instrument rapid exchange system. The scene illustrates the physical workspace configuration, featuring the G1 humanoid robot positioned in front of the custom instrument repository. The background monitor displays the real-time robot FPV captured by the robot's head-mounted visual sensors.

## 4.Result

### 4.1 Time Efficiency Evaluation

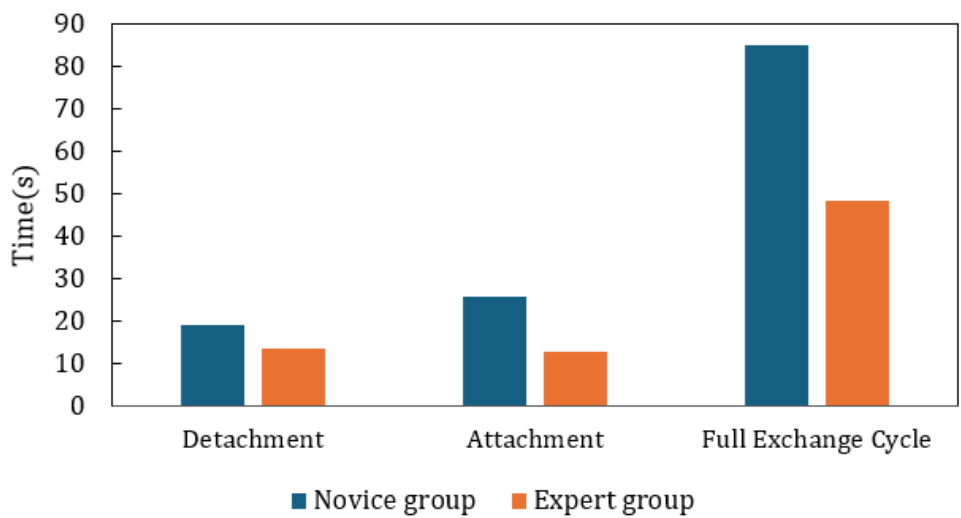

Figure 5. Comparison of time efficiency between novice and expert groups.

To comprehensively evaluate the operational performance of the proposed system, this study assigned the expert group and the novice group to perform 20 trials each of single instrument attachment, single instrument detachment, and a complete exchange cycle. The experimental results indicate that the average time for the expert group to complete a full exchange cycle was 48 seconds, whereas the novice group required 98 seconds. A retrospective analysis of the experimental video recordings revealed that the significant additional time consumed by the novice group was primarily concentrated in the phases of teleoperating the humanoid robot for large-scale spatial movements and fine-tuning the terminal pose alignment.

**4.2 Success Rate and Failure Mode Analysis**

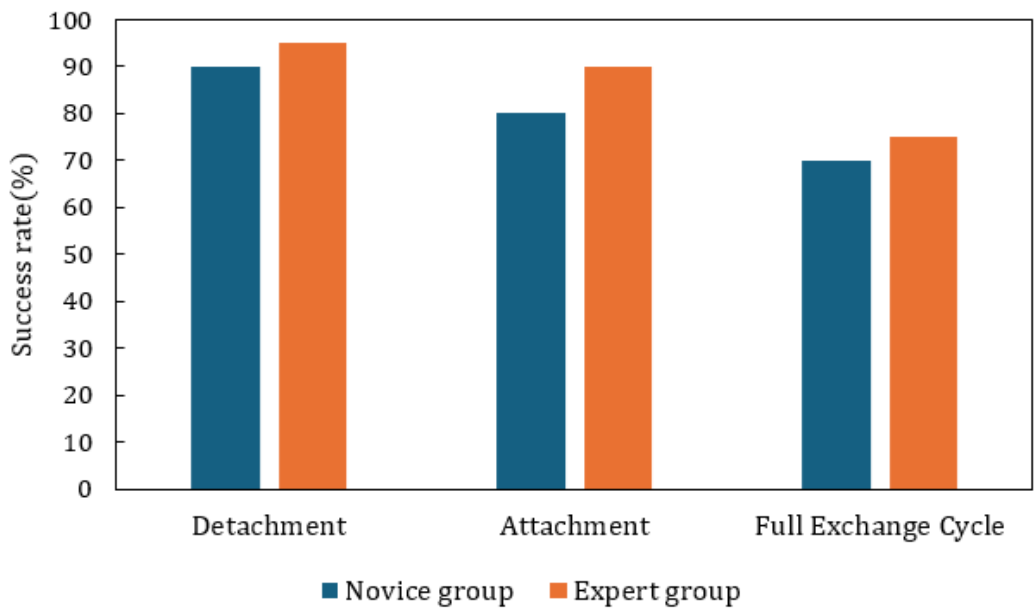

Figure 6. Operational success rates across different task modules.

The operational success rates of the two groups of subjects across various tasks are illustrated in Figure 6; given that the system employs a direct teleoperation mode, subjects could ultimately complete the instrument exchange through continuous trial and error if given unrestricted operational time. Therefore, this study strictly defined a failure as a situation where the operator failed to smoothly complete the docking or releasing action in a single continuous operational flow, thereby necessitating a retraction and retry.

Further failure mode analysis indicates that failures during the detachment process primarily originated from the operator's inability to maintain a horizontal pose at the instrument terminal. An inclined insertion angle prevented the instrument from effectively triggering the limit switch and the auxiliary release mechanism on the base, which subsequently resulted in a failure to activate the rapid exchange button. To address this issue, subsequent hardware iterations will focus on increasing the geometric tolerance of the instrument repository's guide slots. In contrast, failures during the attachment process were mostly attributed to significant axial misalignment between the robotic arm's terminal connection plate and the instrument repository interface. Forcible feeding under this condition caused rigid collisions between the active and passive sides, generating substantial reaction forces; this not only disrupted the force balance of the humanoid robot's bipedal stance, leading to base slippage, but in severe cases, it even accidentally pushed standby instruments in adjacent docking bays out of their slots.

**4.3 Learning Curve and System Limitations**

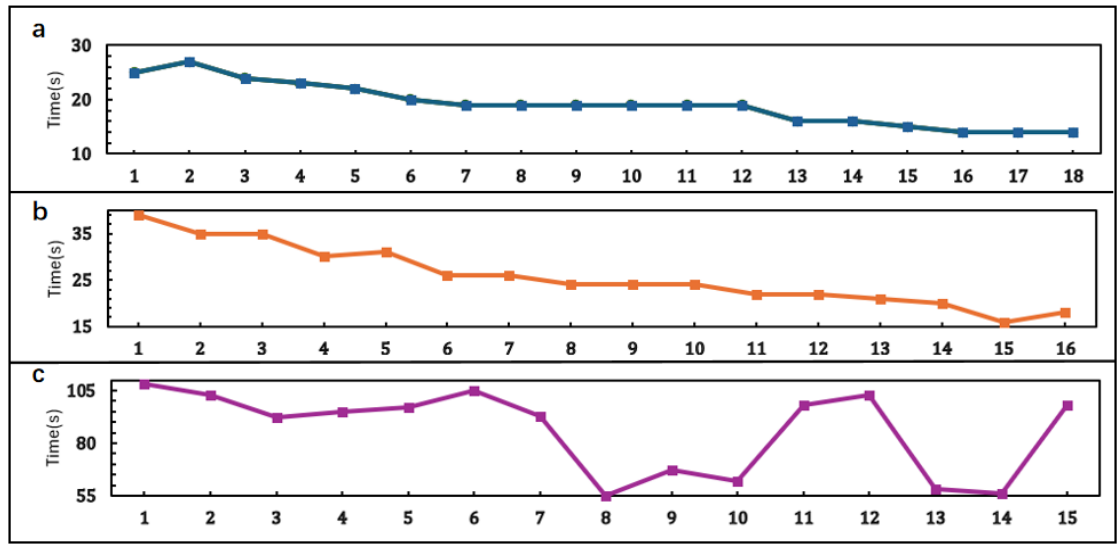

Figure 7. Learning curves of the novice group across successful trials. The line charts track the time consumption variations for the successful attempts out of the 20 planned trials. (a) Time consumption for the single instrument detachment phase. (b) Time consumption for the single instrument attachment phase. (c) Time consumption for the complete exchange cycle.

The time consumption variations of the novice group over 20 consecutive complete operational procedures are shown in Figure 7. The data demonstrate that in the localized instrument attachment and detachment phases, the novice group exhibited a pronounced and rapidly converging learning curve, with their operational time decreasing substantially following short-term training. However, during the macroscopic positioning phase involving the overall spatial movement of the robot, the time cost did not exhibit a clear convergence pattern as the number of training iterations increased. This phenomenon objectively reflects that, at the current stage, teleoperating a high-degree-of-freedom humanoid robot via controllers for long-distance, precise spatial movements still suffers from significant human-robot collaborative instability. This result also highlights the urgent need for automated retrofitting of the accompanying facilities: future research will focus on introducing a smart instrument repository with active addressing or adaptive tracking capabilities, aiming to eliminate the need for large-scale torso movements by the robot, thereby achieving more efficient and stable instrument exchanges while maintaining a static base state.

## 5. Discussion

In recent years, both traditional dedicated surgical platforms and general-purpose humanoid robots have been accelerating their progression toward autonomous operations. Achieving fully automated instrument exchange throughout the entire procedure is a prerequisite for realizing truly autonomous surgery. Although the current stage of this research primarily relies on immersive teleoperation to execute rapid instrument exchange, it constitutes a crucial cornerstone on the path to full autonomy. Leveraging the high-fidelity 3D visual feedback and intuitive spatial motion mapping provided by this system, researchers can safely and efficiently collect massive amounts of high-quality rapid exchange and docking data demonstrated by human experts. These valuable expert operational trajectories will provide indispensable core data support for introducing imitation learning or reinforcement learning algorithms in the future to train humanoid robots to achieve autonomous instrument exchange.

However, the proposed system still possesses certain engineering limitations at the current stage. The physical capacity of the current static instrument repository is limited; further increasing the number of mounted instruments would inevitably force the humanoid robot to undergo larger-scale base movements and pose adjustments when selecting different instruments. As indicated by the experimental results, such macroscopic spatial movements are the primary root cause of increased overall system time consumption and operational instability. Therefore, future research will primarily focus on two dimensions: on the one hand, exploring the system architecture of an active instrument repository, enabling it to autonomously address and present the target instrument according to surgical needs, thereby effectively eliminating the redundant time overhead caused by the robot's body movements; on the other hand, deepening the intelligent control strategies of humanoid robots to facilitate a smooth transition of the system from pure teleoperation to semi-autonomous or fully autonomous rapid exchange, with the ultimate goal of fundamentally reducing the collaborative cognitive load and operational pressure on the primary surgeon during complex minimally invasive surgical procedures.

## 6. Conclusion

This paper proposes a novel immersive teleoperated rapid instrument exchange system for humanoid robots in minimally invasive surgery. We constructed an intuitive hand-eye coordinated control architecture by integrating HMD-based 3D visual feedback with a dedicated physical mechanism that utilizes direct axial-insertion auto-locking and repository-mediated collaborative release. Experimental evaluations confirm the system's high operational robustness and a fast-converging learning curve during localized instrument docking, validating its technical feasibility in constrained clinical spaces. Although long-distance macroscopic spatial teleoperation currently poses stability challenges like base slippage, these highlight clear directions for future development. Subsequent research will introduce an active intelligent instrument repository to eliminate redundant robot movements. Furthermore, the expert trajectories collected via this workflow will provide essential data for imitation and reinforcement learning, ultimately paving the way for autonomous rapid exchange and fully autonomous surgery.